\documentclass[11pt,a4paper]{article}
\usepackage{times,latexsym}
\usepackage{url}
\usepackage[T1]{fontenc}
\usepackage{multirow}
\usepackage{graphicx}
\usepackage{paralist} % compactitem
\usepackage{amsfonts}
\usepackage{amsmath}
\usepackage{amsthm}
\usepackage{makecell}
\usepackage{enumitem}
\usepackage[normalem]{ulem} % uwave
\usepackage{wasysym} % emotion

\usepackage[acceptedWithA]{tacl2018v2}

\usepackage{xspace,mfirstuc,tabulary,multirow}

\newif\iftaclinstructions
\taclinstructionsfalse % AUTHORS: do NOT set this to true
\iftaclinstructions
\renewcommand{\confidential}{}
\renewcommand{\anonsubtext}{(No author info supplied here, for consistency with
TACL-submission anonymization requirements)}
\newcommand{\instr}
\fi

\iftaclpubformat % this "if" is set by the choice of options

\else

\fi

\title{
What You Say and How You Say it: Joint Modeling of Topics and Discourse in Microblog Conversations
}

\author{
 Jichuan Zeng$^1$\Thanks{This work was partially conducted in Jichuan Zeng's internship in Tencent AI Lab. Corresponding author: Jing Li.
 }, Jing Li$^{2*}$, Yulan He$^3$, Cuiyun Gao$^1$, Michael R. Lyu$^1$, Irwin King$^1$   \\
 $^1$Department of Computer Science and Engineering\\
 The Chinese University of Hong Kong, HKSAR, China \\
$^2$Tencent AI Lab, Shenzhen, China \\
$^3$Department of Computer Science, University of Warwick, UK \\
  {$^1$\sf \{jczeng, cygao, lyu, king\}@cse.cuhk.edu.hk}
  \\
  {$^2$\sf ameliajli@tencent.com},  {$^3$\sf yulan.he@warwick.ac.uk}\\
}

\date{}

\begin{document}
\maketitle
\begin{abstract}
This paper presents an \emph{unsupervised} framework for jointly modeling topic content and discourse behavior in microblog conversations. 
Concretely, we propose a \emph{neural} model to discover word clusters indicating what a conversation concerns (i.e., \emph{topics}) and those reflecting how participants voice their opinions (i.e., \emph{discourse}).\footnote{Our datasets and code are available at: \url{http://github.com/zengjichuan/Topic\_Disc}}
Extensive experiments show that our model can yield both coherent topics and meaningful discourse behavior. 
Further study shows that our topic and discourse representations can benefit the classification of microblog messages, especially when they are jointly trained with the classifier.
\end{abstract}

\section{Introduction}

The last decade has witnessed the revolution of communication, where the 
``kitchen table conversations" have been expanded to public discussions on online platforms.
As a consequence, in our daily life, the exposure to new information and the exchange of personal opinions have been mediated through microblogs, one popular online platform genre~\citep{bakshy2015exposure}.
The flourish of microblogs has also led to the sheer quantity of user-created conversations emerging every day, exposing individuals to superfluous information.
Facing such unprecedented number of conversations 
relative to limited attention of individuals,
how shall we automatically extract the critical points and make sense of these microblog conversations?

Towards key focus understanding of a conversation, previous work has shown the benefits of discourse structure~\cite{DBLP:conf/acl/LiLGHW16,DBLP:conf/acl/QinWK17,DBLP:journals/cl/LiSWW2018}, which shapes how messages interact with each other forming the discussion flow and can usefully reflect salient topics raised in the discussion process. 
After all, the topical content of a message naturally occurs in context of the conversation discourse and hence should not be modeled in isolation. 
On the other way around, the extracted topics can reveal the purpose of participants and further facilitate the understanding of their discourse behavior~\cite{DBLP:conf/acl/QinWK17}.
Further, the joint effects of topics and discourse have shown useful to better understand microblog conversations, such as a downstream task to predict user engagements~\cite{DBLP:conf/naacl/ZengLWBSW18}. 

\begin{figure}\small
    \centering
    \scalebox{0.9}{
    \begin{tabular}{|m{7.8cm}|}
    \hline
    ...\\
   M$_1$ $[$\textit{Statement}$]$: Just watched \textbf{HRC} openly endorse a \textbf{gun-control measure} which will fail in front of the \textbf{Supreme Court}. This is a train wreck.\\
   M$_2$ $[$\textit{Comment}$]$: People said the same thing about \textbf{Obama}, and nothing took place.  \textbf{Gun laws} just aren't being enforced like they should be.  :/\\
    M$_3$ $[$\textit{Question}$]$: Okay, hold up. What do you think I'm referencing here? It's not what you're talking about.\\
    M$_4$ $[$\textit{Agreement}$]$: Thought it was about \textbf{gun control}. I'm in agreement that \textbf{gun rights} shouldn't be stripped.\\
    ...\\
    \hline
    \end{tabular}
    }
    %\vskip -0.5em
    \caption{A Twitter conversation snippet about the gun control issue in U.S. \textbf{Topic words} reflecting the conversation focus are in boldface. The \textit{italic} words in $[]$ are our interpretations of the messages' discourse roles. %\jichuan{Do we need to explain that the discourse words are non-topic words?} 
    }\label{fig:example}
   % \vskip -1em
\end{figure}

To illustrate how the topics and discourse interplay in a  conversation, Figure \ref{fig:example} displays a snippet of Twitter conversation. 
As can be seen, the content words reflecting the discussion topics (such as ``\textit{supreme court}'' and ``\textit{gun rights}'') appear in context of the discourse flow, where participants carry the conversation forward via making a statement, giving a comment, asking a question, and so forth. 
Motivated by such an observation, we assume that \emph{a microblog conversation can be decomposed into two crucially different components: one for topical content and the other for discourse behavior}.
Here, the topic components indicate what a conversation is centered around and reflect the important discussion points put forward in the conversation process.
The discourse components signal the \textbf{discourse roles} of messages, such as making a statement, asking a question, and other dialogue acts~\cite{DBLP:conf/naacl/RitterCD10,DBLP:conf/ijcai/JotyCL11}, which further shape the discourse structure of a conversation.\footnote{In this paper, the discourse role refers to a certain type of dialogue act (e.g., \textit{statement} or \textit{question}) for each message. And the discourse structure refers to some combination of discourse roles in a conversation.} 
To distinguish the above two components, we examine the conversation contexts and identify two types of words: \textbf{topic words}, indicating what a conversation focuses on, and \textbf{discourse words}, reflecting how the opinion is voiced in each message. 
For example, in Figure \ref{fig:example}, the topic words ``\textit{gun}'' and ``\textit{control}'' indicate the conversation topic while the discourse word ``\textit{what}'' and ``\textit{?}'' signal the question in M$_3$. 

Concretely, we propose a neural framework built upon topic, enabling the joint exploration of word clusters to represent topic and discourse in microblog conversations. 
Different from the prior models trained on annotated data~\cite{DBLP:conf/acl/LiLGHW16,DBLP:conf/acl/QinWK17}, our model is fully unsupervised, not dependent on annotations for 
either topics or discourse, which ensures its immediate %general 
applicability in any %on some specific 
domain or language. 
Moreover, taking advantages of the recent advances in neural topic models~\cite{DBLP:conf/iclr/SrivastavaC17,DBLP:conf/icml/MiaoGB17}, we are able to approximate Bayesian variational inference without requiring model-specific derivations, while most existing work~\cite{DBLP:conf/naacl/RitterCD10,DBLP:conf/ijcai/JotyCL11,DBLP:conf/icwsm/Alvarez-MelisS16,DBLP:conf/naacl/ZengLWBSW18,DBLP:journals/cl/LiSWW2018} require expertise involved to customize model inference algorithms.
In addition, our neural nature enables end-to-end training of topic and discourse representation learning with other neural models for diverse tasks. 

For model evaluation, we conduct an extensive empirical study on two large-scale Twitter datasets. 
The intrinsic results show that our model can produce  latent topics and discourse roles with better interpretability than the state-of-the-art models from previous studies. 
The extrinsic evaluations on a tweet classification task exhibit the model's ability to capture useful representations for microblog messages.
Particularly, our model enables an easy combination with existing neural models for end-to-end training, such as CNN, which is shown to perform better in classification than the pipeline approach without joint training.   
\section{Related Work}

Our work is in the line with previous studies that employ \emph{non-neural} models to leverage discourse structure for extracting topical content from conversations~\cite{DBLP:conf/acl/LiLGHW16,DBLP:conf/acl/QinWK17,DBLP:journals/cl/LiSWW2018}. 
\citet{DBLP:conf/naacl/ZengLWBSW18} explores how discourse and topics jointly affect user engagements in microblog discussions.
Different from them, we build our model in a \emph{neural network} framework, where the joint effects of topic and discourse representations can be exploited for various downstream deep learning tasks in an end-to-end manner. In addition, we are inspired by the prior research that only models topics or conversation discourse. In the following, we discuss them in turn.

\paragraph{Topic Modeling.} Our work is closely related with the topic model studies. In this field, despite of the huge success achieved by the springboard topic models (e.g., pLSA~\cite{DBLP:conf/sigir/Hofmann99} and LDA~\cite{DBLP:conf/nips/BleiNJ01}), and their extensions~\cite{DBLP:conf/nips/BleiGJT03,DBLP:conf/uai/Rosen-ZviGSS04}, the applications of these models have been limited to formal and well-edited documents, such as news reports~\cite{DBLP:conf/nips/BleiGJT03} and scientific articles~\cite{DBLP:conf/uai/Rosen-ZviGSS04}, attributed to their reliance on document-level word collocations.
When processing short texts, such as the messages on microblogs, it is likely that the performance of these models will be inevitably compromised, due to the severe data sparsity issue.

To deal with such an issue, many previous efforts incorporate the \emph{external} representations, such as word embeddings~\cite{DBLP:journals/tacl/NguyenBDJ15,DBLP:conf/sigir/LiWZSM16,DBLP:conf/sigir/ShiLJSL17} and knowledge~\cite{DBLP:conf/ijcai/SongWWLC11,DBLP:conf/emnlp/YangDB15,DBLP:conf/ijcai/HuLSXN16}, pre-trained on large-scale high-quality resources. 
Different from them, our model learns topic and discourse representations only with the internal data and thus can be widely applied on scenarios where the specific external resource is unavailable.

In another line of the research, most prior work focuses on how to enrich the context of short messages. 
To this end, biterm topic model (BTM)~\cite{DBLP:conf/www/YanGLC13} extends a message into a biterm set with all combinations of any two distinct words appearing in the message.
On the contrary, our model allows the richer context in a conversation to be exploited, where word collocation patterns can be captured beyond a short message. 
 
In addition, there are many methods employing some heuristic rules to aggregate short messages into long pseudo-documents, such as those based on authorship~\cite{DBLP:conf/kdd/Hong010,DBLP:conf/ecir/ZhaoJWHLYL11} and hashtags~\cite{DBLP:conf/icwsm/RamageDL10,DBLP:conf/sigir/MehrotraSBX13}. Compared with these methods, we model messages in context of their conversations, which has been demonstrated to be a more natural and effective text aggregation strategy for topic modeling~\cite{DBLP:conf/icwsm/Alvarez-MelisS16}.  

\paragraph{Conversation Discourse.} Our work is also in the area of discourse analysis for conversations, ranging from the prediction of the shallow discourse roles on utterance level \cite{J00-3003,DBLP:conf/naacl/JiHE16,DBLP:conf/acl/ZhaoLE18} to the discourse parsing for a more complex conversation structure~\cite{DBLP:conf/acl/ElsnerC08,DBLP:journals/coling/ElsnerC10,DBLP:conf/emnlp/AfantenosKAP15}.
In this area, most existing models heavily rely on the data annotated with discourse labels for learning~\cite{DBLP:conf/acl/ZhaoZE17}.
Different from them, our model, in a fully unsupervised way,
identifies distributional word clusters to represent latent discourse factors in conversations.
Although such latent discourse variables have been studied in previous work~\cite{DBLP:conf/naacl/RitterCD10,DBLP:conf/ijcai/JotyCL11,DBLP:conf/naacl/JiHE16,DBLP:conf/acl/ZhaoLE18}, none of them explores the effects of latent discourse on the identification of conversation topic, which is a gap our work fills in.

\section{Our Neural Model for Topics and Discourse in Conversations}\label{sec:approach}

This section introduces our neural model that jointly explores latent representations for topics and discourse in conversations. We first present an overview of our model in Section~\ref{ssec:overview}, followed by the model generative process and inference procedure in Section~\ref{ssec:generation} and \ref{ssec:inference}, respectively.  

\subsection{Model Overview} \label{ssec:overview}

In general, our model aims to learn coherent word clusters that reflect the latent topics and discourse roles embedded in the microblog conversations. 
To this end, we distinguish two latent components in the given collection: \emph{topics} and \emph{discourse}, each represented by a certain type of word distribution (distributional word cluster). 
Specifically, at the corpus level, we assume there are $K$ topics, represented by $\phi^T_k$ ($k=1,2,\dots, K$), and $D$ discourse roles, captured with $\phi^D_d$ ($d=1,2,\dots, D$).
$\phi^T$ and $\phi^D$ are all multinomial word distributions over the vocabulary size $V$.
Inspired by the neural topic models in \citet{DBLP:conf/icml/MiaoGB17}, our model encodes topic and discourse distributions ($\phi^T$ and $\phi^D$) as latent variables in a neural network and learns the parameters via back propagation. 

\begin{figure}[t]
	\centering
	\includegraphics[width=0.48 \textwidth]{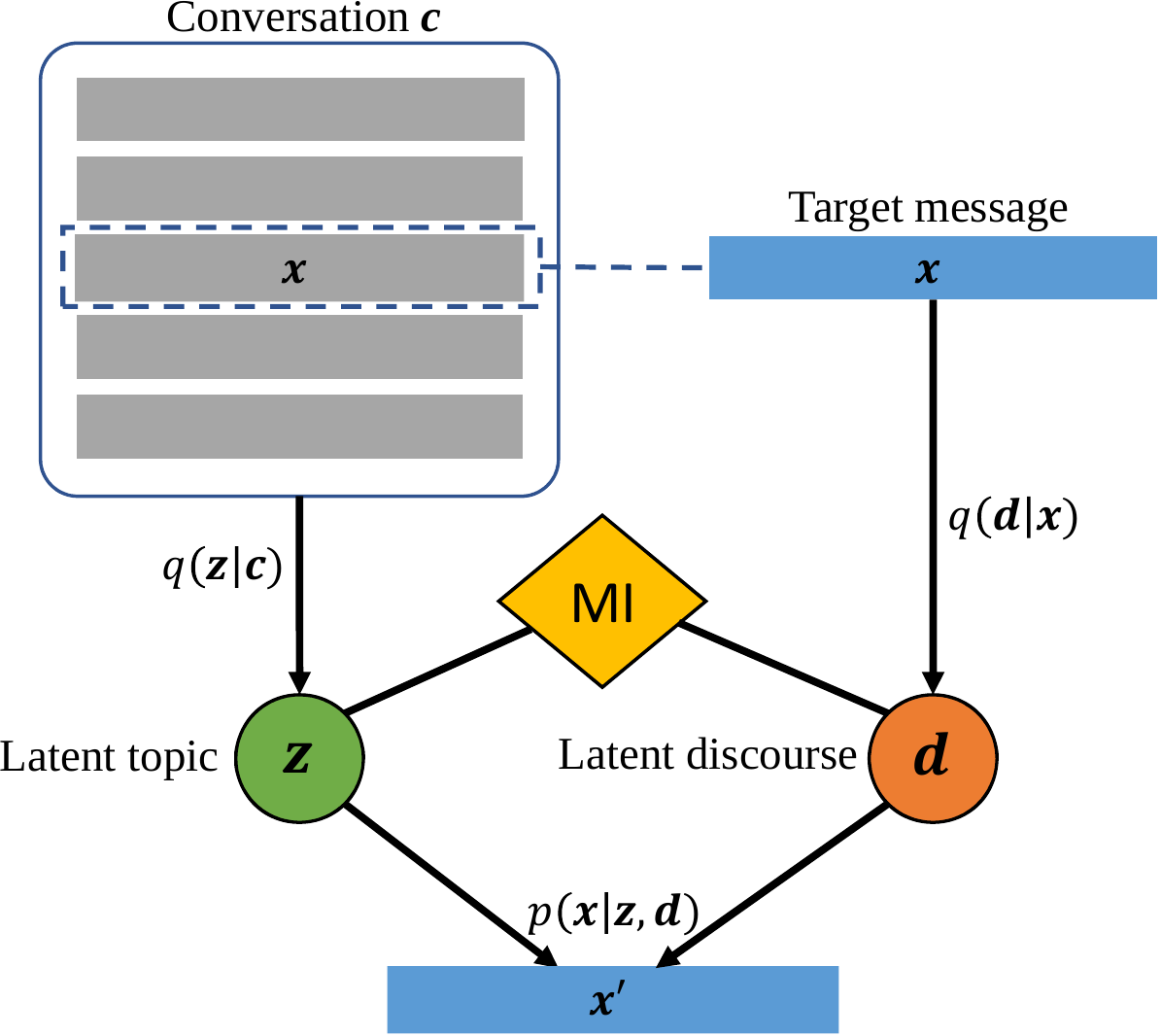}
	%\vskip -0.5em
	\caption{
    The architecture of our neural framework that jointly models latent topics and latent discourse. 
    }
	\label{fig:framework}
    %\vskip -1em
\end{figure}

Before touching the details of our model, we first describe how we formulate the input. 
On microblogs, as a message might have multiple replies, messages in an entire conversation can be organized as a tree with replying relations~\cite{DBLP:conf/acl/LiLGHW16,DBLP:journals/cl/LiSWW2018}.
Though the recent progress in recursive models allows the representation learning from the tree-structured data, previous studies have pointed out that, in practice, sequence models serve as a more simple yet robust alternative~\cite{DBLP:conf/emnlp/LiLJH15}. 
In this work, we follow the common practice in most conversation modeling research~\cite{DBLP:conf/naacl/RitterCD10,DBLP:conf/ijcai/JotyCL11,DBLP:conf/acl/ZhaoLE18} to take a conversation as a sequence of turns. 
To this end, each conversation tree is flattened into root-to-leaf paths. Each one of such paths is hence considered as a conversation instance, and a message on the path corresponds to a conversation turn~\cite {DBLP:conf/sigdial/ZarishevaS15,DBLP:conf/coling/CerisaraJOL18,DBLP:conf/www/JiaoL0M18}.

The overall architecture of our model is shown in Figure~\ref{fig:framework}. Formally, we formulate a conversation $\bf c$ as a sequence of messages $( {\bf x}_1, {\bf x}_2,\dots,{\bf x}_{M_c})$, where $M_c$ denotes the number of messages in $\bf c$. 
In the conversation, each message ${\bf x}$, named as the \textbf{target message}, is fed into our model sequentially. 
Here we process the target message ${\bf x}$ as the bag-of-words (BoW) term vector ${\bf x}_{BoW} \in \mathbb{R}^V$, following the bag-of-words assumption in most topic models~\cite{DBLP:conf/nips/BleiGJT03,DBLP:conf/icml/MiaoGB17}. The conversation, $\bf c$, where the target message ${\bf x}$ is involved, is considered as the context of ${\bf x}$. It is also encoded %Together with ${\bf x}$, we also input its conversation $\bf c$ 
in the BoW form (denoted as ${\bf c}_{BoW} \in \mathbb{R}^V$) and fed into our model. In doing so, we ensure context of the target message is incorporated %to be encoded 
while learning its latent representations. 
%of the target message.

Following the previous practice in neural topic models~\cite{DBLP:conf/icml/MiaoGB17,DBLP:conf/iclr/SrivastavaC17}, we employ the variational auto-encoder (VAE)~\cite{DBLP:journals/corr/KingmaW13} to resemble the data generative process via two steps. 
First, given the target message ${\bf x}$ and its conversation $\bf c$, our model converts them into two latent variables: topic variable $\bf z$ and discourse variable $\bf d$. 
Then, using the intermediate representations captured by $\bf z$ and $\bf d$, we reconstruct the target message, ${\bf x}'$.

\subsection{Generative Process}\label{ssec:generation}

In this section, we first describe the two latent variables in our model: the topic variable $\bf z$ and the discourse variable $\bf d$. Then, we present our data generative process from the latent variables.

\paragraph{Latent Topics.}
For latent topic learning, we examine the main discussion points in the context of a conversation.
Our assumption is that messages in the same conversation tend to focus on similar topics~\cite{DBLP:journals/cl/LiSWW2018,DBLP:conf/naacl/ZengLWBSW18}. Concretely, we define the latent topic variable ${\bf z}\in \mathbb{R}^K$ at the \emph{conversation} level and generate the topic mixture of $\bf c$, denoted as a $K$-dimentional distribution $\theta$, via a softmax construction conditioned on ${\bf z}$~\cite{DBLP:conf/icml/MiaoGB17}.

\paragraph{Latent Discourse.} 
For modeling the discourse structure of conversations, we capture the \emph{message}-level discourse roles reflecting the dialogue acts of each message, as is done in~\citet{DBLP:conf/naacl/RitterCD10}. 
Concretely, given the target message $\bf x$, we employ a $D$-dimensional one-hot vector to represent the latent discourse variable $\bf d$, where the high bit indicates the index of a discourse word distribution that can best express $\bf x$'s discourse role. In the generative process, the latent discourse $\bf d$ is drawn from a multinomial distribution with parameters estimated from the input data.

\paragraph{Data Generative Process}
As mentioned previously, our entire framework is based on VAE, which consists of an encoder and a decoder. The encoder maps a given input into latent topic and discourse representations and the decoder reconstructs the original input from the latent representations. In the following, we first describe the decoder followed by the encoder. 

In general, our \emph{decoder} is learned to reconstruct the words in the target message $\bf x$ (in the BoW form) from the latent topic $\bf z$ and latent discourse $\bf d$. We show the generative story that reflects the reconstruction process below:

\begin{compactitem}
\item Draw the latent topic  $\bf z\sim\mathcal{N}(\boldsymbol{\mu}, \boldsymbol{\sigma}^2)$ 
\item $\bf c$'s topic mixture $\theta = \operatorname{softmax}(f_\theta ({\bf z}))$
% \begin{compactitem}
% \item For each message $m=1$ to $M_c$:
%\begin{compactitem}
\item Draw the latent discourse  ${\bf d} \sim Multi(\boldsymbol{\pi})$
%\item $\phi = \operatorname{softmax}(f_\phi ({\bf \theta}))$\jichuan{RM}
\item For the $n$-th word in $\bf x$
\begin{compactitem}
\item $\beta_n = \operatorname{softmax}(f_{\phi^T} ({\bf \theta})+f_{\phi^D} ({\bf d}))$
\item Draw the word $w_{n}\sim Multi(\beta_{n})$
%w_{n}\sim p(w_{n}\,|\,\boldsymbol{W}, \theta)$
%\end{compactitem}
%\end{compactitem}
\end{compactitem}
\end{compactitem}
where $f_*(\cdot)$ is a neural perceptron, with a linear transformation of inputs activated by a non-linear transformation. Here we use rectified linear units (ReLUs)~\cite{DBLP:conf/icml/NairH10} as the activate functions. In particular, the weight matrix of $f_{\phi^T}(\cdot)$ (after the softmax normalization) is considered as the topic-word distributions $\phi^T$. The discourse-word distributions $\phi^D$ are similarly obtained from $f_{\phi^D}(\cdot)$.

For the \emph{encoder}, we learn the parameters $\boldsymbol{\mu}$, $\boldsymbol{\sigma}$, and $\boldsymbol{\pi}$ from the input ${\bf x}_{BoW}$  and ${\bf c}_{BoW}$ (the BoW form of the target message and its conversation), following the formula below: %and define their generation functions as following:
\begin{equation}\small
\begin{aligned}
\boldsymbol{\mu} = f_{\mu}&(f_e({\bf c}_{BoW}))
,\, \log\boldsymbol{\sigma} = f_{\sigma}(f_e({{\bf c}_{BoW}}))
\\
&\boldsymbol{\pi} = \operatorname{softmax}(f_{\pi}({\bf x}_{BoW}))
\end{aligned}
\end{equation}

\subsection{Model Inference}\label{ssec:inference}

For the objective function of our entire framework, we take three aspects into account: the learning of latent topics and discourse, the reconstruction of the target messages, and the separation %distinguishing 
of topic-associated words and discourse-related words. 

\paragraph{Learning Latent Topics and Discourse.} For learning the latent topics/discourse %parameters 
in our model, we employ the variational inference~\cite{DBLP:journals/corr/BleiKM16} to approximate posterior distribution over the latent topic $\bf z$ and the latent discourse $\bf d$ given all the training data. To this end, we maximize the variational lower bound $\mathcal{L}_{z}$ for $\bf z$ and $\mathcal{L}_{d}$ for $\bf d$, each defined as following:

\begin{equation}\small
\begin{aligned}\label{eq:topic-discourse}
&\mathcal{L}_{z} = \mathbb{E}_{q({{\bf z}\,|\,{\bf c}})}[p({\bf c}\,|\,{\bf z})] - D_{KL}(q({\bf z}\,|\,{\bf c})\,||\,p({\bf z}))
\\
&\mathcal{L}_{d} = \mathbb{E}_{q({{\bf d}\,|\,{\bf x}})}[p({\bf x}\,|\,{\bf d})] - D_{KL}(q({\bf d}\,|\,{\bf x})\,||\,p({\bf d}))
\end{aligned}
\end{equation}

\noindent $q(\bf z\,|\,{\bf c})$ and $q({\bf d}\,|\,{\bf x})$ are approximated posterior probabilities describing how the latent topic $\bf z$ and the latent discourse $\bf d$ are generated from the data. 
$p({\bf c}\,|\,{\bf z})$ and $p({\bf x}\,|\,{\bf d})$ represent the corpus likelihoods conditioned on the latent variables.
Here to facilitate coherent topic production, in $p({\bf c}\,|\,{\bf z})$, we penalize stop words' likelihood to be generated from latent topics following~\citet{DBLP:journals/cl/LiSWW2018}.
$p({\bf z})$ follows the standard normal prior $\mathcal{N}({\bf 0}, {\bf I})$ and $p({\bf d})$ is the uniform distribution $Unif(0,1)$.
$D_{KL}$ refers to the Kullback-Leibler divergence (KLD) that ensures the approximated posteriors to be close to the true ones.
For more derivation details, we refer the readers to \citet{DBLP:conf/icml/MiaoGB17}.

\paragraph{Reconstructing target messages.} 
From the latent variables $\bf z$ and $\bf d$, the goal of our model is to reconstruct the target message $\bf x$. The corresponding learning objective is to maximize $\mathcal{L}_{x}$ defined as:

\begin{equation}\small
\begin{aligned}
&\mathcal{L}_{x} = \mathbb{E}_{q({z\,|\,{\bf x}})q(d\,|\,{\bf c})}[\log p({\bf x}\,|\,{\bf z},{\bf d})]
\end{aligned}
\end{equation}

\noindent Here we design $\mathcal{L}_{x}$ to ensure that the learned latent topics and discourse can reconstruct $\bf x$.

\paragraph{Distinguishing Topics and Discourse.} 
Our model aims to distinguish word distributions for topics ($\phi^T$) and discourse ($\phi^D$), which enables topics and discourse to capture different information in conversations. 
Concretely, we employ the mutual information, given below, to 
measure the mutual dependency between the latent topics $\bf z$ and the latent discourse $\bf d$.
\footnote{The distributions in Eq. \ref{eq:mi-general} are all conditional probability distributions given the target message $\bf x$ and its conversation $\bf c$. We omit the conditions for simplicity.}

\begin{equation}\small\label{eq:mi-general}
\begin{aligned}
\mathbb{E}_{q({\bf z})q({\bf d})}[\log \frac{p({\bf z},{\bf d})}{p({\bf z})p({\bf d})}]
\end{aligned}
\end{equation}

\noindent Eq. \ref{eq:mi-general} can be further derived as the Kullback-Leibler divergence of the conditional distribution, $p({\bf d}\,|\,{\bf z})$, and marginal distribution, $p({\bf d})$. The derived formula, defined as the mutual information loss ($\mathcal{L}_{MI}$) and shown in Eq. \ref{eq:mi-kl}, is used to map $\bf z$ and $\bf d$ into the separated semantic space.

\begin{equation}\small\label{eq:mi-kl}
\begin{aligned}
\mathcal{L}_{MI} = \mathbb{E}_{q({\bf z})}[D_{KL}(p({\bf d}\,|\,{\bf z}) || p({\bf d}))]
\end{aligned}
\end{equation}

\noindent We can hence minimize $\mathcal{L}_{MI}$ for guiding our model to separate word distributions that represent topics and discourse.

\paragraph{The Final Objective.} To capture the joint effects of the learning objectives described above ($\mathcal{L}_{z}$, $\mathcal{L}_{d}$, $\mathcal{L}_{x}$, and $\mathcal{L}_{MI}$), we design the final objective function for our entire framework as following:

\begin{equation}\label{eq: final-obj}\small
\begin{aligned}
\mathcal{L} = \mathcal{L}_{z} + \mathcal{L}_{d} + \mathcal{L}_{x} - \lambda \mathcal{L}_{MI} 
\end{aligned}
\end{equation}

\noindent where the hyperparameter $\lambda$ is the trade-off parameter for balancing between the MI loss ($\mathcal{L}_{MI}$) and the other learning objectives. 
By maximizing the final objective $\mathcal{L}$ via back propagation, the word distributions of topics and discourse can be jointly learned from microblog conversations.\footnote{To smooth the gradients in implementation, for ${\bf z}\sim\mathcal{N}(\boldsymbol{\mu},\boldsymbol{\sigma})$, we apply the reparameterization on $\bf z$ ~\cite{DBLP:journals/corr/KingmaW13,DBLP:conf/icml/RezendeMW14}, and for ${\bf d}\sim Multi(\boldsymbol{\pi})$, we adopt the Gumbel-Softmax trick~\cite{DBLP:journals/corr/MaddisonMT16,DBLP:journals/corr/JangGP16}.}

\begin{table}[]\footnotesize
    \centering
    \begin{tabular}{|l|rrrr|}
         \hline
         
         \multirow{2}{*}{\textbf{Datasets}}& \# of &  Avg msgs & Avg words&\multirow{2}{*}{|Vocab|}\\
         &convs&per conv&per msg&\\
         \hline
         TREC & 116,612 & 3.95 & 11.38 & 9,463\\
         TWT16 & 29,502 & 8.67 & 14.70 & 7,544\\
         \hline
    \end{tabular}
    %\vskip -0.5em
    \caption{Statistics of the two datasets containing Twitter conversations.}\label{tab:statistics}
   %\vskip -1em
\end{table}

\section{Experimental Setup}

\paragraph{Data Collection.}
For our experiments, we collected two microblog conversation datasets from Twitter. 
One is released by the TREC 2011 microblog track (henceforth \textbf{TREC}), containing conversations concerning a wide rage of topics.\footnote{\url{http://trec.nist.gov/data/tweets/}} 
The other is crawled from January to June 2016 with Twitter streaming API\footnote{\url{https://developer.twitter.com/en/docs/tweets/filter-realtime/api-reference/post-statuses-filter.html}} (henceforth \textbf{TWT16}, short for Twitter 2016), following the way of building TREC dataset.
During this period, there are a large volume of discussions centered around U.S. presidential election. 
In addition, for both datasets, we apply Twitter search API\footnote{\url{https://developer.twitter.com/en/docs/tweets/search/api-reference/get-saved searches-show-id.}} to retrieve the missing tweets in the conversation history, as the Twitter streaming API (used to collect both datasets) only returns sampled tweets from the entire pool. 

The statistics of the two experiment datasets are shown in Table \ref{tab:statistics}. For model training and evaluation, we randomly sampled $80\%$, $10\%$, and $10\%$ of the data to form the training, development, and test set, respectively.

\paragraph{Data Preprocessing.} We preprocessed the data 
with the following steps. 
First, non-English tweets were filtered out. Then, hashtags, mentions (@username), and links were replaced with generic tags ``HASH'', ``MENT'', and ``URL'', respectively. Next, the natural languge toolkit (NLTK) was applied for tweet tokenization.\footnote{\url{https://www.nltk.org/}} After that, all letters were normalized to lower cases. Finally, words occurred less than $20$ times were filtered out from the data.

\paragraph{Parameter Setting.}
To ensure comparable results with  \citet{DBLP:journals/cl/LiSWW2018} (the prior work focusing on the same task as ours), in the topic coherence evaluation, 
we follow their setup to report the results under two sets of $K$ (the number of topics): $K=50$ and $K=100$, and with the number of discourse roles ($D$) set to $10$. The analysis for the effects of $K$ and $D$ will be further presented in Section \ref{ssec:further-discussion}. 
For all the other hyper-parameters, we tuned them on development set by grid search. The trade-off parameter $\lambda$ (defined in Eq. \ref{eq: final-obj}), balancing the MI loss and the other objective functions, is set to $0.01$. In model training, we use Adam optimizer~\cite{DBLP:journals/corr/KingmaB14} and run $100$ epochs with early stop strategy adopted. 

\paragraph{Baselines.} In topic modeling experiments, we consider the five topic model baselines treating each tweet as a document: LDA~\cite{DBLP:conf/nips/BleiGJT03}, BTM~\cite{DBLP:conf/www/YanGLC13}, LF-LDA, LF-DMM~\cite{DBLP:journals/tacl/NguyenBDJ15}, and NTM~\cite{DBLP:conf/icml/MiaoGB17}.
In particular, BTM and LF-DMM are the state-of-the-art topic models for short texts. BTM explores the topics of all word pairs (biterms) in each message to alleviate data sparsity in short texts. LF-DMM incorporates word embeddings pre-trained on external data to expand semantic meanings of words, so does LF-LDA. 
In \citet{DBLP:journals/tacl/NguyenBDJ15}, LF-DMM, based on one-topic-per-document Dirichlet Multinomial Mixture (DMM)~\cite{DBLP:journals/ml/NigamMTM00}, was reported to perform better than LF-LDA, based on LDA. For LF-LDA and LF-DMM, we use GloVe Twitter embeddings~\cite{DBLP:conf/emnlp/PenningtonSM14} as the pre-trained word embeddings.\footnote{\url{https://nlp.stanford.edu/projects/glove/}}

For the discourse modeling experiments, we compare our results with  LAED~\cite{DBLP:conf/acl/ZhaoLE18}, a VAE-based representation learning model for conversation discourse. %, into account for comparison. 
In addition, for both topic and discourse evaluation, we compare with \citet{DBLP:journals/cl/LiSWW2018}, a recently proposed model for microblog conversations, where topics and discourse are jointly explored with a \emph{non-neural} framework. Besides the existing models from previous studies, we also compare with the variants of our model that only models topics (henceforth \textsc{Topic only}) or discourse (henceforth \textsc{Disc only}).\footnote{
In our ablation without mutual information loss ($\mathcal{L}_{MI}$ defined in Eq. \ref{eq:mi-general}), topics and discourse are learned independently. 
Thus, its topic representation can be used for the output of \textsc{Topic only}, so does its discourse one for \textsc{Disc only}.
} 
Our joint model of topics and discourse is referred to as \textsc{Topic+Disc}. 

In the preprocessing process for the baselines, we removed stop words and punctuation for topic models unable to learn discourse representations following the common practice in previous work~\cite{DBLP:conf/www/YanGLC13,DBLP:conf/icml/MiaoGB17}. For the other models, stop words and punctuation were retained in the vocabulary considering their usefulness as discourse indicators~\cite{DBLP:journals/cl/LiSWW2018}.

\section{Experimental Results}\label{sec:result}

In this section, we first report the 
topic coherence results in Section~\ref{ssec:coherece}, followed by a discussion in Section~\ref{ssec:discourse} comparing the latent discourse roles discovered by our model 
with the manually annotated dialogue acts. 
Then, we study whether we can capture useful representations for microblog messages in a tweet classification task (in Section \ref{ssec:classification}).
A qualitative analysis, showing some example topics and discourse roles, is further provided in Section~\ref{ssec:qualitative-analysis}. Finally, in Section~\ref{ssec:further-discussion}, we provide more discussions on our model.

\subsection{Topic Coherence}\label{ssec:coherece}

For the topic coherence, we adopt the $C_v$ scores measured via the open-source Palmetto toolkit as our evaluation metric.\footnote{\url{https://github.com/dice-group/Palmetto}}
$C_v$ scores assume that the top $N$ words in a coherent topics (ranked by likelihood) tend to co-occur in the same document and have shown comparable evaluation results to human judgments~\cite{DBLP:conf/wsdm/RoderBH15}. 
Table~\ref{tab:topic} shows the average $C_v$ scores over the produced topics given $N=5$ and $N=10$. The values range from $0.0$ to $1.0$ and higher scores indicate better topic coherence. We can observe that:

\begin{table}[t]\small
	\center
	\scalebox{0.9}{\begin{tabular}{|l|rr|rr|}
		\hline
		%\multicolumn{2}{|c|}{\multirow{2}{*}{Models}}
		\multirow{2}{*}{\textbf{Models}} & \multicolumn{2}{c|}{$K=50$} & \multicolumn{2}{c|}{$K=100$} \\
		\cline{2-5}
		& TREC & TWT16 & TREC & TWT16 \\
		\hline
		\hline
		%\multirow{3}{*}{Traditional topic models} & 
		\underline{\textbf{Baselines}} &&&&\\
		LDA & 0.467 & 0.454 & 0.467 & 0.454 \\
		BTM & 0.460 & 0.461 & 0.466 & 0.463 \\
	    LF-DMM
	    & 0.456 & 0.448 & 0.463 & 0.466 \\
	    LF-LDA & 0.470 & 0.456& 0.467& 0.453\\
		%\hline
		%Nueral topic model & 
		NTM & 0.478 & 0.479 & 0.482 & 0.443 \\
		%\hline
		%Non-neural joint training model& 
		\citet{DBLP:journals/cl/LiSWW2018} & 0.463 & 0.433 & 0.464 & 0.435 \\
		\hline
		\hline
		%Nueral joint training model &
		\underline{\textbf{Our models}} &&&&\\
	   \textsc{Topic only} & 0.478 & 0.482 & 0.481 & 0.471 \\
		\textsc{Topic+Disc} & \textbf{0.485} & \textbf{0.487} & \textbf{0.496} & \textbf{0.480} \\
		\hline
	\end{tabular}
}
\caption{$C_v$ coherence scores for latent topics produced by different models. The best result in each column is highlighted in \textbf{bold}. Our joint model \textsc{Topic+Disc} achieves significantly better coherence scores than all the baselines ($p<0.01$, paired test).} %Our \textsc{Topic+Disc} performs the best.}
%\vskip -1em
\label{tab:topic}
\end{table}

\vspace{0.5em}

\noindent$\bullet$~\textit{\textbf{Models assuming a single topic for each message  %One-topic-per-post models 
do not work well.}} 
It has long been pointed out that the one-topic-per-message assumption (each message contains only one topic) helps topic models alleviate the data sparsity issue in short texts on microblogs~\cite{DBLP:conf/ecir/ZhaoJWHLYL11,DBLP:conf/ijcai/QuanKGP15,DBLP:journals/tacl/NguyenBDJ15,DBLP:journals/cl/LiSWW2018}. However, we observe contradictory results since both LF-DMM and \citet{DBLP:journals/cl/LiSWW2018}, following this assumption, achieve generally worse performance than the other models. 
%The opposite observations 
This might be attributed to the large-scale data used in our experiments (each dataset has over $250$K messages as shown in Table \ref{tab:statistics}), which potentially provide richer word co-occurrence patterns and thus partially alleviate the data sparsity issue.

\noindent$\bullet$~\textit{\textbf{Pre-trained word embeddings do not bring benefits.}} Comparing LF-LDA with LDA, we found that they give similar coherence scores. This shows that with sufficiently large training data, with or without using the pre-trained word embeddings do not make any difference in the topic coherence results.

\noindent$\bullet$~\textit{\textbf{Neural models perform better than non-neural baselines.}} %are effective in topic modeling.}} 
When comparing the results of neural models (\textsc{NTM} and our models) with the other baselines, we find the former yield topics with better coherence scores in most cases. 

\noindent$\bullet$~\textit{\textbf{Modeling topics in conversations is effective.}} Among neural models, we found our models outperform NTM (without exploiting conversation contexts). This shows that the conversations provide useful context and enables more coherent topics to be extracted from the entire conversation thread instead of a single short message. 

\noindent$\bullet$~\textit{\textbf{Modeling topics together with discourse helps produce more coherent topics.}} %From the results from our two models, 
We can observe better results with the joint model \textsc{Topic+Disc} in comparison with the variant considering topics only. This shows that \textsc{Topic+Disc}, via the joint modeling of topic- and discourse-word distributions (reflecting non-topic information), can better separate topical words from non-topical ones, hence resulting in more coherent topics.  

\begin{table}[t]\small
	\center
\begin{tabular}{|l|r|r|r|}
		\hline
		\textbf{Models} & Purity & Homogeneity & VI\\
		\hline
		\hline
		\underline{\textbf{Baselines}}&&&\\
		LAED & 0.505 & 0.022 & 6.418\\
		\citet{DBLP:journals/cl/LiSWW2018} & 0.511 & 0.096 & 5.540 \\
		\hline
		\hline
        \underline{\textbf{Our models}}&&&\\
		\textsc{Disc only} & 0.510 & 0.112 & 5.532 \\
		\textsc{Topic+Disc} & \textbf{0.521} & \textbf{0.142} & \textbf{5.097} \\
		\hline
	\end{tabular}
	%\vspace{-0.5em}
\caption{
%Comparison of the quality of generated discourse clusters in terms of  purity and homogeneity.
The purity, homogeneity, and variation of information (VI) scores for the latent discourse roles measured against the human-annotated dialogue acts. For purity and homogeneity, higher scores indicate better performance, while for VI scores, lower is better. In each column, the best results are in \textbf{boldface}. Our joint model \textsc{Topic+Disc} significantly outperforms all the baselines ($p<0.01$, paired t-test).
}\label{tab:purity}
%\vspace{-0.5em}
\end{table}

\subsection{Discourse Interpretability}\label{ssec:discourse}

In this section, we evaluate whether our model can discover meaningful discourse representations. To this end, we train the comparison models for discourse modeling on the TREC dataset and test the learned latent discourse on a benchmark dataset released by \citet{DBLP:conf/coling/CerisaraJOL18}. 
The benchmark dataset consists of $2,217$ microblog messages forming $505$ conversations collected from Mastodon\footnote{\url{https://mastodon.social}}, a microblog platform exhibiting Twitter-like user behavior~\cite{DBLP:conf/coling/CerisaraJOL18}. 
For each message, there is a human-assigned discourse label, selected from one of the $15$ dialogue acts, such as \textit{question}, \textit{answer}, \textit{disagreement}, etc.

For discourse evaluation, we measure whether the model-produced discourse assignments are consistent with the human-annotated dialogue acts. 
Hence following \citet{DBLP:conf/acl/ZhaoLE18}, we assume that an interpretable latent discourse role should cluster messages labeled with the same dialogue act. 
Therefore, we adopt  purity~\cite{DBLP:books/daglib/0021593}, homogeneity~\cite{DBLP:conf/emnlp/RosenbergH07}, and variation of information (VI)~\cite{DBLP:conf/colt/Meila03,DBLP:conf/acl/GoldwaterG07} as our automatic evaluation metrics. 
Here, we set $D=15$ to ensure the number of latent discourse roles to be the same as the number of manually-labeled dialogue acts. 
Table \ref{tab:purity} shows the comparison results of the average scores over the $15$ latent discourse roles. Higher values indicate better performance for purity and homogeneity, while for VI, lower is better. 

It can be observed that our models exhibit generally better performance, showing the effectiveness of our framework in inducing interpretable discourse roles. 
Particularly, we observe the best results achieved by our joint model \textsc{Topic+Disc}, which 
is learned to distinguish topic- and discourse-words, important in recognizing indicative words to reflect latent discourse. 

\begin{figure}[t]
	\centering
% 	\includegraphics[width=0.45 \textwidth]{figures/cluster_vis.pdf}
% 	\caption{
%     Visualization of the alignment of predicted clusters and true labels. The x-axis is the predicted clusters (from 1 to 14 here), the y-axis is the true labels from~\cite{DBLP:conf/coling/CerisaraJOL18}. Each character along with y-axis represents a dialogue act type.
    
%     \footnotesize{
%     \scalebox{0.9}{
%     \begin{tabular}{|c|c||c|c|}
%     \hline 
%     CHR  & Act & CHR & Act  \\ \hline
%     D  & disagreement & S & suggest  \\
%     A & agreement & Q & yes/no question \\
%     O & wh*/open question & W & open+choice answer \\
%     H & initial greetings & T & thanking \\
%     R & request & M & sympathy \\
%     V & explicit performance & J & exclamation \\
%     F & acknowledge & E & offer\\
%     \hline
%     \end{tabular}
%     }}
%     }
%\vskip -1em
    \begin{minipage}{7.5cm}
     \includegraphics[width=7.0cm]{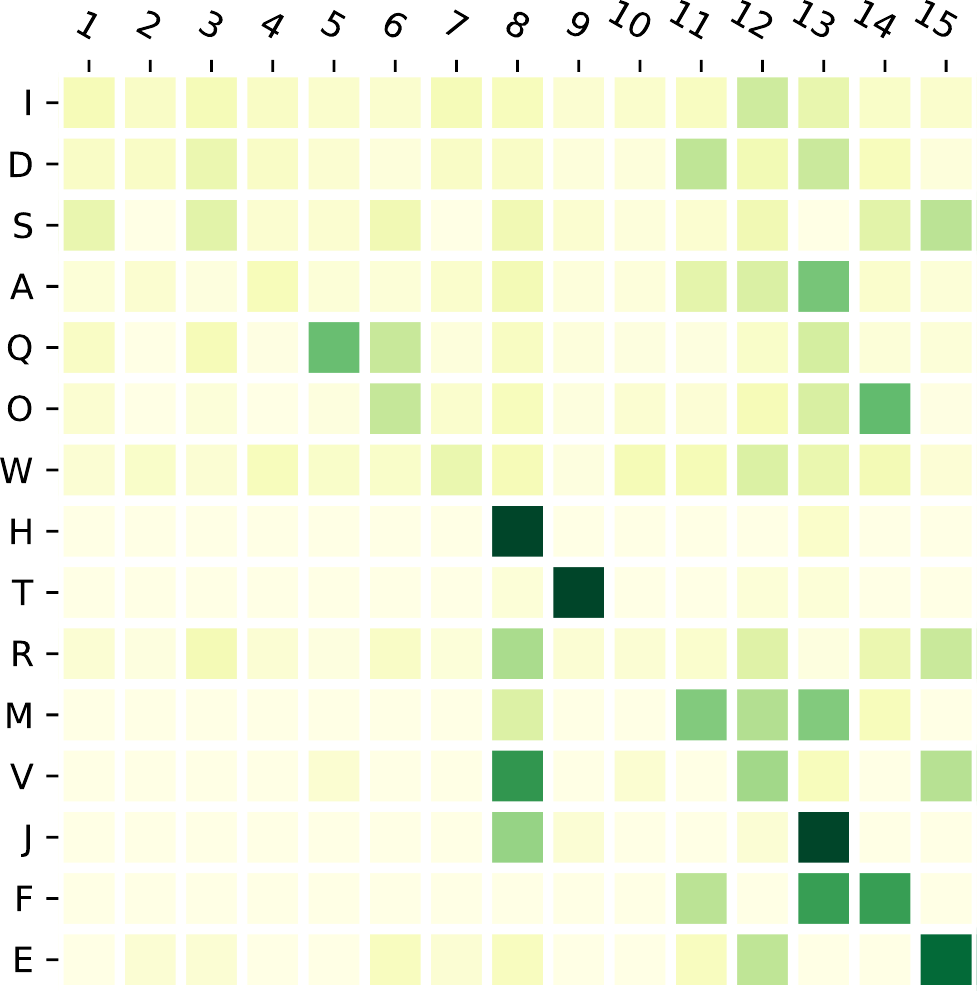}
     
     {\footnotesize \textsf{I}: statement, \textsf{D}: disagreement, \textsf{S}: suggest, \textsf{A}: agreement, \textsf{Q}: yes/no question, \textsf{O}: wh*/open question, \textsf{W}: open+choice answer, \textsf{H}: initial greetings, \textsf{T}: thanking, \textsf{R}: request, \textsf{M}: sympathy, \textsf{V}: explicit performance, \textsf{J}: exclamation, \textsf{F}: acknowledge, and \textsf{E}: offer.\par}
    \end{minipage}
    \caption{
    A heatmap showing the alignments of the latent discourse roles and human-annotated dialogue act labels. 
    Each line visualizes the distribution of messages with the corresponding dialogue act label over varying discourse roles (indexed from 1 to 15), where darker colors indicate higher values.
    % The x-axis is the predicted clusters (from 1 to 14 here), the y-axis is the true labels from~\cite{DBLP:conf/coling/CerisaraJOL18}. Each character along with y-axis represents a dialogue act type, with details shown below the heatmap.
    }
    %\vskip -1em
    % \footnote{Each character along with y-axis represents a dialogue act type, where D: disagreement, S: suggest, A: agreement, Q: yes/no question, O: wh*/open question, W: open+choice answer, H: initial greetings, T: thanking, R: request, M: sympathy, V: explicit performance, J: exclamation, F: acknowledge, and E: offer.}
	\label{fig:disc-align}
\end{figure}

To further analyze the consistency of varying latent discourse roles (produced by our \textsc{Topic+Disc} model) with the human-labeled dialogue acts, Figure \ref{fig:disc-align} displays a heatmap, where each line visualizes how the messages  with a dialogue act distribute over varying discourse roles. 
It is seen that among all dialogue acts, our model discovers more interpretable latent discourse for ``\textit{greetings}'', ``\textit{thanking}'', ``\textit{exclamation}'', and ``\textit{offer}'', where most messages are clustered into one or two dominant discourse roles. It may be because these dialogue acts can be relatively easier to detect based on their associated indicative words, %better indicated by the observed words, 
such as the word ``\textit{thanks}'' for ``\textit{thanking}'', and the word ``\textit{wow}'' for ``\textit{exclamation}''. 

\subsection{Message Representations}\label{ssec:classification}
To further evaluate our ability to capture effective representations for microblog messages, we take tweet classification as an example and test the classification performance with the topic and discourse representations as features. 
\begin{table}[t]\small
	\center
	\scalebox{0.96}{\begin{tabular}{|l|r|r|r|r|}
		\hline
		\multirow{2}{*}{\textbf{Models}}& \multicolumn{2}{c|}{TREC} & \multicolumn{2}{c|}{TWT16} \\
		\cline{2-5}
		& Acc & Avg F1 & Acc & Avg F1 \\
		\hline
		\hline
		\underline{\textbf{Baselines}} &&&&\\
		BoW & 0.120 & 0.026 & 0.132 & 0.030 \\
		TF-IDF & 0.116 & 0.024 & 0.153 & 0.041 \\
		LDA & 0.128 & 0.041 & 0.146 & 0.046 \\
		BTM & 0.123 & 0.035 & 0.167 & 0.054 \\
		LF-DMM & 0.158 & 0.072 & 0.162 & 0.052 \\
		NTM & 0.138 & 0.042 & 0.186 & 0.068 \\
		\hline
		Our model & \textbf{0.259} & \textbf{0.180} & \textbf{0.341} & \textbf{0.269} \\
		\hline
	\end{tabular}
}
%\vskip -0.5em
\caption{Evaluation of tweet classification results in accuracy (Acc) and average F1 (Avg F1). 
%where the user-generated hashtags are considered as the class labels. 
Representations learned by various models serve as the classification features. For our model, both the topic and discourse representations are fed into the classifier.
}\label{tab:topic_svm}
%\vskip -0.5em
\end{table}
Here the user-generated hashtags capturing the topics of online messages are used as the proxy class labels~\cite{DBLP:conf/acl/LiLGHW16,DBLP:conf/emnlp/ZengLSGLK2018}. 
We construct the classification dataset from TREC and TWT16 with the following steps.
First, we removed the tweets without hashtags.
Second, we ranked hashtags by their frequencies.
Third, we manually removed the hashtags that are not topic-related (e.g. ``\textit{\#fb}'' for indicating the source of tweets from Facebook), and combined the hashtags referring to the same topic (e.g. ``\textit{\#DonaldTrump}'' and ``\textit{\#Trump}'').
Finally, we selected the top $50$ frequent hashtags, and all tweets containing these hashtags as our classification dataset. 
Here, we simply use the support vector machines (SVMs) as the classifier, since our focus is to compare the representations learned by various models. \citet{DBLP:journals/cl/LiSWW2018} is unable to produce vector representation on tweet level, hence not considered here.

Table~\ref{tab:topic_svm} shows the classification results of accuracy and average F1 on the two datasets with the representations learned by various models serving as the classification features.
We observe that our model outperforms other models with a large margin. The possible reasons are two folds. 
First, our model derives topics from conversation threads and thus potentially yields better message representations.
Second, the discourse representations (only produced by our model) are indicative features for hashtags, because users will exhibit various discourse behaviors in discussing diverse topics (hashtags). For instance, we observe prominent ``\textit{argument}'' discourse from tweets with ``\textit{\#Trump}'' and ``\textit{\#Hillary}'', attributed to the controversial opinions to the two candidates in the 2016 U.S. presidential election.

\subsection{Example Topics and Discourse Roles}\label{ssec:qualitative-analysis}

We have shown that joint modeling of topics and discourse presents superior performance on quantitative measure. 
In this section, we qualitatively analyze the interpretability of our outputs via analyzing the word distributions of some example topics and discourse roles. 

\begin{table}\footnotesize
\centering
\scalebox{1.0}{\begin{tabular}{|l|m{4.8cm}|}
\hline
LDA &  \textcolor{blue}{\uwave{people}} trump police violence gun death protest guns \textcolor{red}{\uline{flag}} shot \\
\hline
BTM &  gun guns \textcolor{blue}{\uwave{people}} police wrong right \textcolor{blue}{\uwave{think}} law agree black \\
\hline
LF-DMM & gun police black \textcolor{blue}{\uwave{said}} \textcolor{blue}{\uwave{people}} guns killing ppl amendment laws \\
\hline
\citet{DBLP:journals/cl/LiSWW2018} & wrong don trump gun \textcolor{blue}{\uwave{understand}} laws agree guns \textcolor{blue}{\uwave{doesn}} \textcolor{blue}{\uwave{make}} \\
\hline
NTM &  gun \textcolor{blue}{\uwave{understand}} \textcolor{blue}{\uwave{yes}} guns world dead \textcolor{blue}{\uwave{real}} discrimination trump silence \\ 
\hline
\hline
\textsc{Topic only} & shootings gun guns cops charges control \textcolor{blue}{\uwave{mass}} commit \textcolor{blue}{\uwave{know}} agreed\\
\hline
\textsc{Topic+Disc} &   guns gun shootings chicago shooting cops firearm criminals commit laws  \\ 
\hline
\end{tabular}
}
%\vskip -0.5em
\caption{Top $10$ representative words of example latent topics discovered from the TWT16 dataset. We interpret the topics as ``gun control'' by the displayed words. \textcolor{blue}{\uwave{Non-topic words}} are wave-underlined and in blue, while \textcolor{red}{\uline{off-topic words}} are underlined and in red.}\label{tab:topic-words}
% \uwave
%\vskip -1em
\end{table}  

\paragraph{Example Topics.}
Table \ref{tab:topic-words} lists the top $10$ words of some example latent topics discovered by various models from the TWT16 dataset. 
According to the words shown, we can interpret the extracted topics as ``gun control'' --- discussion about gun law and the failure of gun control in Chicago. 
We observe that LDA wrongly includes off-topic word ``\textit{flag}''. 
From the outputs of BTM, LF-DMM, \citet{DBLP:journals/cl/LiSWW2018}, and our \textsc{Topic only} variant, though we do not find off-topic words, there are some non-topic words, such as ``\textit{said}'' and ``\textit{understand}''.\footnote{Non-topic words do not clearly indicate the corresponding topic, while off-topic words are more likely to appear in other topics.} 
The output of our \textsc{Topic+Disc} model appears to be the most coherent, with words such as ``\textit{firearm}'' and ``\textit{criminals}'' included, which are clearly relevant to ``gun control''. 
Such results indicate the benefit of examining the conversation contexts and jointly exploring topics and discourse in them.

\paragraph{Example Discourse Roles.}
To qualitatively analyze whether our \textsc{Topic+Disc} model can discover interpretable discourse roles, we select the top $10$ words from the distributions of some example discourse roles and list them in Table \ref{tab:disc_sample}. 
It can be observed that there are some meaningful word clusters reflecting varying discourse roles found without any supervision.
Interestingly, we observe that the latent discourse roles from TREC and TWT16, though learned separately, exhibit some notable overlap in their associated top 10 words, particularly for ``\textit{question}'' and ``\textit{statement}''.
We also note that ``\textit{argument}'' is represented by very different words. The reason is that TWT16 contains a large volume of arguments centered around candidate Clinton and Trump, resulting in the frequent appearance of words like ``\textit{he}'' and ``\textit{she}''.

\begin{table}[t]\small
	\center
	\scalebox{1.0}{
	\begin{tabular}{c}
    \centering
	\includegraphics[width=0.46\textwidth]{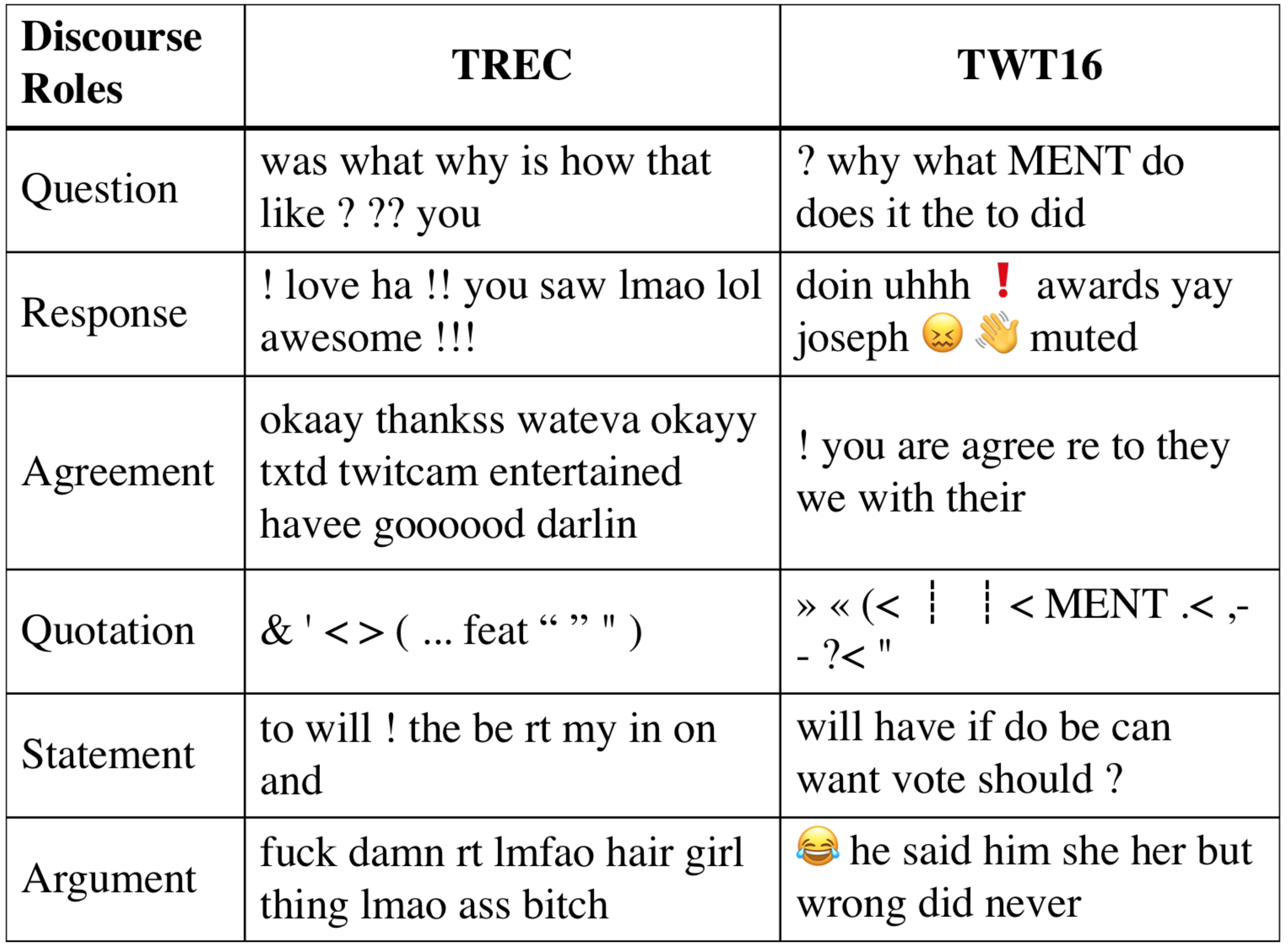}
	\end{tabular}
}
\caption{Top 10 representative words of example discourse roles learned from TREC and TWT16. The discourse roles of the word clusters are manually assigned according to their associated words. }\label{tab:disc_sample}
\end{table}

\subsection{Further Discussions}\label{ssec:further-discussion}

In this section, we further present more discussions on our joint model: \textsc{Topic+Disc} .

% \begin{figure}[h]
% \begin{tabular}{cc}
% \includegraphics[width=0.23\textwidth]{figures//topic_num.png} & \includegraphics[width=0.23\textwidth]{figures//disc_num.png}\\

% \\

% \end{tabular}
% \caption{Parameter Study.}
% \label{fig:parameter}
% \end{figure}
\begin{figure}[t]
	\centering	\includegraphics[width=0.48 \textwidth]{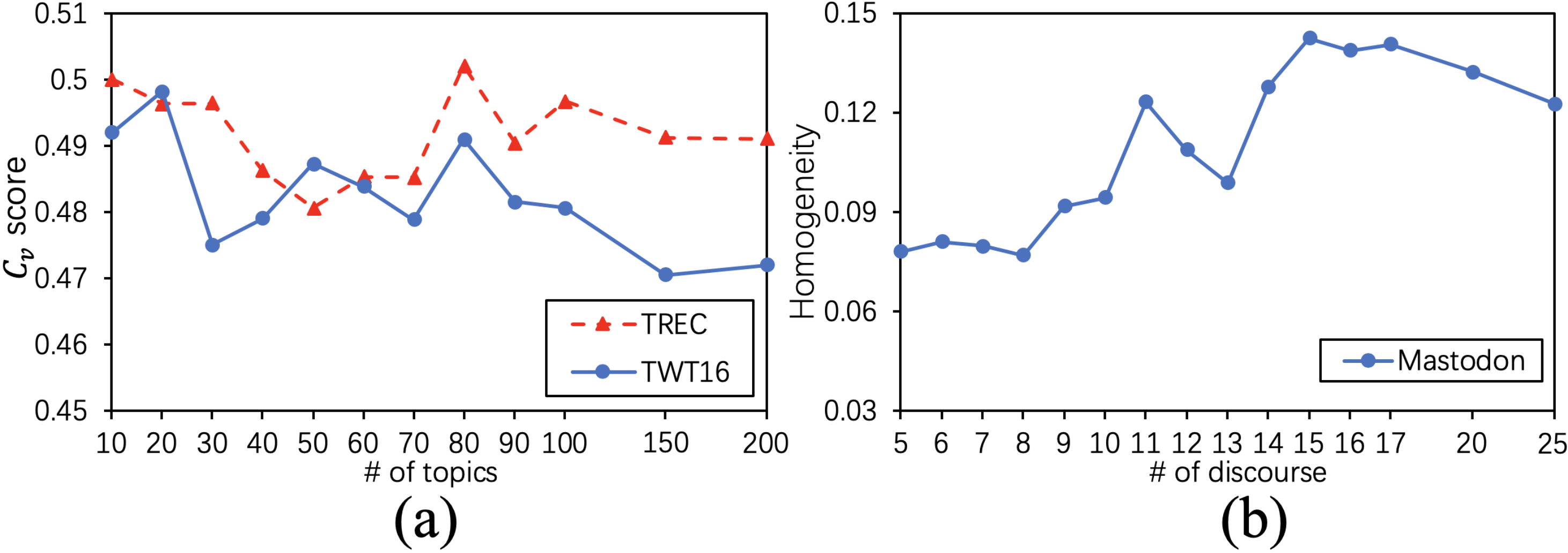}
    %\vskip -0.5em
    \caption{(a) The impact of topic numbers. The horizontal axis: the number of topics; The vertical axis: the $C_v$ topic coherence. (b) The impact of discourse numbers. The horizontal axis: the number of discourse; The vertical axis: the homogeneity measure.}
    %\vskip -1em
\label{fig:topic_num}
\end{figure}

\paragraph{Parameter Analysis.} Here we study the two important hyper-parameters in our model, the number of topics ($K$) and the number of discourse roles ($D$). 
In Figure \ref{fig:topic_num}, we show the $C_v$ topic coherence given varying $K$ in (a) and the homogeneity measure given varying $D$ in (b). 
As can be seen, the curves corresponding to the performance on topics and discourse are not monotonic. 
In particular, better topic coherence scores are achieved given relatively larger topic numbers for TREC with the best result observed at $K=80$. On the contrary, the optimum topic number for TWT16 is $K=20$, while increasing the number of topics results in worse $C_v$ scores in general.
This may be attributed to the relatively centralized topic concerning U.S. election in the TWT16 corpus.
For discourse homogeneity, the best result is achieved given $D=15$, with same the number of manually annotated dialogue acts in the benchmark.

\begin{figure}[t]
	\centering
	\includegraphics[width=0.48 \textwidth]{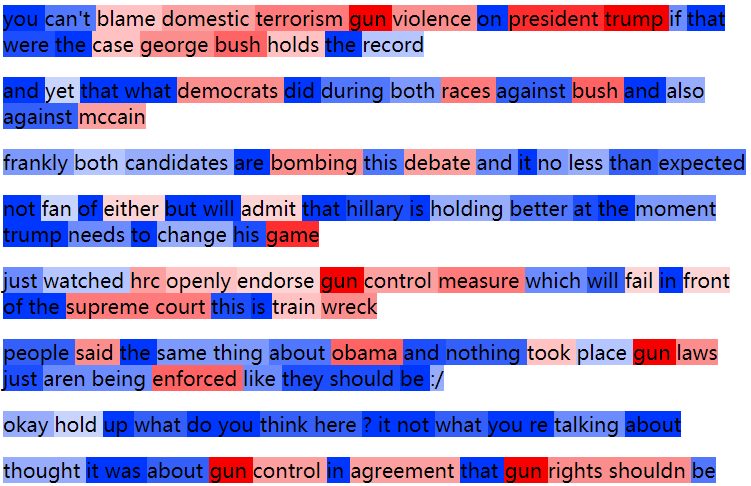}
	%\vskip -0.5em
	\caption{
    Visualization of the topic-discourse assignment of a twitter conversion from TWT16. The annotated blue words are pone to be discourse words, and the red are topic words. The shade is indicating the confidence of current assignment.
    }
	\label{fig:case-disc}
    %\vskip -1.0em
\end{figure}

\paragraph{Case Study.}
To further understand why our model learns meaningful representations for topics and discourse, we present a case study based on the example conversation shown in Figure \ref{fig:example}. Specifically, we visualize the topic words (with $p(w\,|\,{\bf z})>p(w\,|\,{\bf d})$) in red and the rest words in blue to indicate discourse. Darker red indicates the higher topic likelihood ($p(w\,|\,{\bf z})$) while darker blue shows the higher discourse likelihood ($p(w\,|\,{\bf d})$). %are defined similarly for discourse.
The results are shown in Figure \ref{fig:case-disc}. 
We can observe that topic and discourse words are well separated by our model, which explains why it can generate high-quality representations for both topics and discourse.

\paragraph{Model Extensibility.}
Recall that in the Introduction, we have mentioned that our neural-based model 
%, in a neural networks framework, 
has an advantage to be easily combined with other neural network architectures and allows for the joint training of both models. %the joint training process with them. 
Here we take message classification (with the setup in Section \ref{ssec:classification}) as an example, and study whether joint training our model with convolutional neural network (CNN)~\cite{DBLP:conf/emnlp/Kim14}, the widely-used
%state-of-the-art 
model on short text classification, can bring benefits to the classification performance.
We set the embedding dimension to $200$, with random initialization.
The results are shown in Table \ref{tab:joint_cls}, where we observe that joint training our model and the classifier can successfully boost the classification performance.

\paragraph{Error Analysis.} We further analyze the errors in our outputs. For topics, taking a closer look at their word distributions, we found that our model sometimes mix sentiment words with topic words. For example, among the top 10 words of a topic ``\textit{win people illegal americans hate lt racism social tax wrong}'', there are words ``\textit{hate}'' and ``\textit{wrong}'', expressing sentiment rather than conveying topic-related information. 
This is due to the prominent co-occurrences of topic words and sentiment words in our data, which results in the similar distributions for topics and sentiment. 
Future work could focus on the further separation of sentiment and topic words.

For discourse, we found %two major errors from the word distributions. First, 
that our model can induce some discourse roles beyond the $15$ manually defined dialogue acts %reflecting the absent dialogue acts 
in the Mastodon dataset \cite{DBLP:conf/coling/CerisaraJOL18}. For example, as shown in Table \ref{tab:disc_sample}, our model discover the ``\textit{quotation}'' discourse from both TREC and TWT16, which is however not defined in the Mastodon dataset. This perhaps should not be considered as an error. We argue that it is not sensible to pre-define a fixed set of dialogue acts for diverse microblog conversations due to the rapid change and a wide variety of user behaviors in social media. Therefore, future work should involve a better alternative to evaluate the latent discourse without relying on manually defined dialogue acts. We also notice that our model sometimes fails to identify discourse behaviors requiring more in-depth semantic understanding, such as sarcasm, irony, and humor. This is because our model detects latent %The reason is that our 
discourse %representations can only be discovered from 
purely based on the observed words, while the detection of sarcasm, irony, or humor %while deep semantic analysis 
requires deeper language understanding, which is beyond the capacity of our model. 

\begin{table}[]\small
	\center
	\scalebox{0.95}{\begin{tabular}{|l|r|r|r|r|}
		\hline
		\multirow{2}{*}{\textbf{Models}}& \multicolumn{2}{c|}{TREC} & \multicolumn{2}{c|}{TWT16} \\
		\cline{2-5}
		& Acc & Avg F1 & Acc & Avg F1 \\
		\hline
		\hline
		CNN only & 0.199 & 0.167 & 0.334 & 0.311 \\
		\hline
		Separate-Train & 0.284 & 0.270 & 0.391 & 0.390 \\
		\hline
		Joint-Train & \textbf{0.297} & \textbf{0.286} & \textbf{0.428} & \textbf{0.413} \\
		\hline
	\end{tabular}
}
%\vskip -0.5em
\caption{Accuracy (Acc) and average F1 (Avg F1) on tweet classification (hashtags as labels).
CNN only: CNN without using our representations. Seperate-Train: CNN fed with our pre-trained representations. Joint-Train: Joint training CNN and our model.}
%\vskip -1em
\label{tab:joint_cls}
\end{table}

\section{Conclusion}

We have presented a neural framework that jointly explores topic and discourse from microblog conversations. 
Our model, in an unsupervised manner, examines the conversation contexts and discovers word distributions that reflect latent topics and discourse roles.
Results from extensive experiments show that our model can generate coherent topics and meaningful discourse roles. In addition, our model can be easily combined with other neural network architectures (such as CNN) and allows for joint training, which has presented better message classification results compared to the pipeline approach without joint training.
\section*{Acknowledgements}
This work is partially supported by the Research Grants Council of the Hong Kong Special Administrative Region, China (No. CUHK 14208815 and No. CUHK 14210717 of the General Research Fund), Innovate UK (grant No. 103652), and Microsoft Research Asia (2018 Microsoft Research Asia Collaborative Research Award). We thank Shuming Shi, Dong Yu, and TACL reviewers and editors for the insightful suggestions on various aspects of this work.

\bibliography{tacl2018}
\bibliographystyle{acl_natbib}

\end{document}